\documentclass[conference]{IEEEtran}
\IEEEoverridecommandlockouts
\usepackage{cite}
\usepackage{amsmath,amssymb,amsfonts}
\usepackage{algorithmic}
\usepackage{graphicx}
\usepackage{subcaption}  
\usepackage{float} 
\usepackage{textcomp}
\usepackage{xcolor}
\usepackage{url}
\usepackage{placeins}
\usepackage{enumitem}

\def\BibTeX{{\rm B\kern-.05em{\sc i\kern-.025em b}\kern-.08em
    T\kern-.1667em\lower.7ex\hbox{E}\kern-.125emX}}
\begin{document}

\title{
A Multimodal Fusion Framework for Bridge Defect Detection with Cross-Verification
\\

\thanks{}
}

\author{\IEEEauthorblockN{Ravi Datta Rachuri}
    \IEEEauthorblockA{
        \textit{School of Computing} \\
        \textit{George Mason University}\\
        Fairfax, VA \\
        rrachuri@gmu.edu}
\and
\IEEEauthorblockN{\textsuperscript{} Duoduo Liao}
\IEEEauthorblockA{\textit{School of Computing} \\
\textit{George Mason University}\\
Fairfax, VA \\
dliao2@gmu.edu}
\and
\IEEEauthorblockN{\textsuperscript{} Samhita Sarikonda}
\IEEEauthorblockA{\textit{School of Computing} \\
\textit{George Mason University}\\
Fairfax, VA \\
ssariko@gmu.edu}
\and
    \IEEEauthorblockN{\textsuperscript{} Datha Vaishnavi Kondur}
\IEEEauthorblockA{\textit{School of Computing} \\
\textit{George Mason University}\\
Fairfax, VA \\
dkondur@gmu.edu}
}
\maketitle

\begin{abstract}

This paper presents a pilot study introducing a multimodal fusion framework for the detection and analysis of bridge defects, integrating Non-Destructive Evaluation (NDE) techniques with advanced image processing to enable precise structural assessment. By combining data from Impact Echo (IE) and Ultrasonic Surface Waves (USW) methods, this preliminary investigation focuses on identifying defect-prone regions within concrete structures, emphasizing critical indicators such as delamination and debonding. Using geospatial analysis with alpha shapes, fusion of defect points, and unified lane boundaries, the proposed framework consolidates disparate data sources to enhance defect localization and facilitate the identification of overlapping defect regions. Cross-verification with adaptive image processing further validates detected defects by aligning their coordinates with visual data, utilizing advanced contour-based mapping and bounding box techniques for precise defect identification. The experimental results, with an F1 score of 0.83, demonstrate the potential efficacy of the approach in improving defect localization, reducing false positives, and enhancing detection accuracy, which provides a foundation for future research and larger-scale validation. This preliminary exploration establishes the framework as a promising tool for efficient bridge health assessment, with implications for proactive structural monitoring and maintenance.

\end{abstract}


\section{Introduction}

Bridges are critical to transportation infrastructure, yet they face progressive deterioration due to aging, environmental exposure, and heavy traffic loads. This deterioration manifests as surface and subsurface defects, such as delamination, debonding, internal cracking, and elasticity changes, which pose significant safety risks if undetected \cite{b5, b11}. Traditional visual inspections are often insufficient for identifying subsurface anomalies and rely heavily on subjective judgment, necessitating advanced, objective assessment methods for accurate structural health evaluations \cite{b2, b12}. Non-Destructive Evaluation (NDE) techniques, such as Impact Echo (IE) and Ultrasonic Surface Waves (USW), have emerged as reliable tools for bridge inspections. IE excels at detecting internal flaws such as voids and cracks, while USW evaluates elasticity and stiffness to identify material degradation. However, their independent use is limited by their specific focus areas \cite{b13, b14, b21}. 

 This pilot study proposes a multimodal data fusion framework integrating IE and USW data to overcome these limitations. Data fusion is the process of integrating multiple data sources to generate a unified representation, leveraging the strengths of each source to provide a more comprehensive understanding of the observed phenomenon \cite{b13, b15}. In this study, data fusion combines results from IE and USW techniques to enhance the reliability and accuracy of structural health assessments. By combining these complementary modalities, data fusion ensures a more holistic assessment of structural anomalies. It corroborates defect-prone regions identified by both techniques, reducing uncertainties and increasing confidence in the detected anomalies \cite{b3, b19}. Furthermore, the integration of multimodal data allows for the identification of overlapping defect areas, which can be crucial for prioritizing maintenance activities and resource allocation \cite{b13, b21}. Geospatial techniques such as Alpha Shape Analysis (ASA) are employed to localize overlapping defect regions, while adaptive image processing validates defect zones through contour-based cross-verification, reducing false positives and enhancing reliability \cite{b21, b22}. 
 
 The proposed framework establishes an efficient approach for sustainable infrastructure monitoring and provides a foundation for future large-scale applications. The primary objectives of this study are: (1) to detect bridge defects, reliably by integrating NDE techniques (IE and USW) into a multimodal data fusion framework; and (2) to implement cross-verification through image processing to validate data-derived defect locations against visual inspection images, ensuring reliability and reducing false positives in detected defects.

The key contributions of this study are outlined as follows:
\begin{itemize}
    \item A unified framework integrating IE and USW data, leveraging their complementary strengths to enhance the detection of both internal and surface defects.
    
    \item A novel approach combining ASA with lane-specific boundary segmentation to achieve precise localization of overlapping defect regions.
    
    \item Adaptive contour-based image processing techniques for validating data-derived defect locations, significantly reducing false positives and ensuring alignment with visual inspection data.
    
    \item The robustness of the proposed multimodal framework through cross-verification of defect detection results between IE and USW datasets, enhancing reliability and detection accuracy.
    
    \item A scalable foundation for large-scale implementation of multimodal defect detection techniques, supporting the sustainability and resilience of transportation infrastructure assessment.
\end{itemize}

\section{Related Work}

\subsection{NDE Modalities - IE and USW}

NDE techniques have been widely used to detect subsurface defects in bridge structures. Among these, IE and USW are prominent methods due to their ability to detect critical subsurface issues such as debonding and delamination\cite{b5, b13}. Previous studies highlight the specific strengths of each modality: IE is adept at identifying internal flaws, including voids and cracks, by analyzing stress wave reflections within concrete \cite{b6}. USW measures material elasticity, identifying areas of reduced stiffness that may indicate material deterioration or potential failure points \cite{b5, b9}. Despite their advantages, each modality has inherent limitations. IE effectively locates voids but is limited in assessing elasticity changes within materials—an area where USW excels. Conversely, USW provides valuable elasticity information but is less precise in localizing specific defects like voids and cracks \cite{b6,b9}. Studies by Gucunski et al. \cite{b5} and Carino et al. \cite{b6} emphasize that while both methods contribute valuable insights individually, their limitations hinder a full understanding of structural health when used alone. This limitation highlights the need for a combined approach using both IE and USW in a complementary manner to provide a holistic diagnostic view of bridge structures \cite{b8,b10}.

\begin{figure*} [h]
    \centering
    \includegraphics[width=0.8\textwidth]{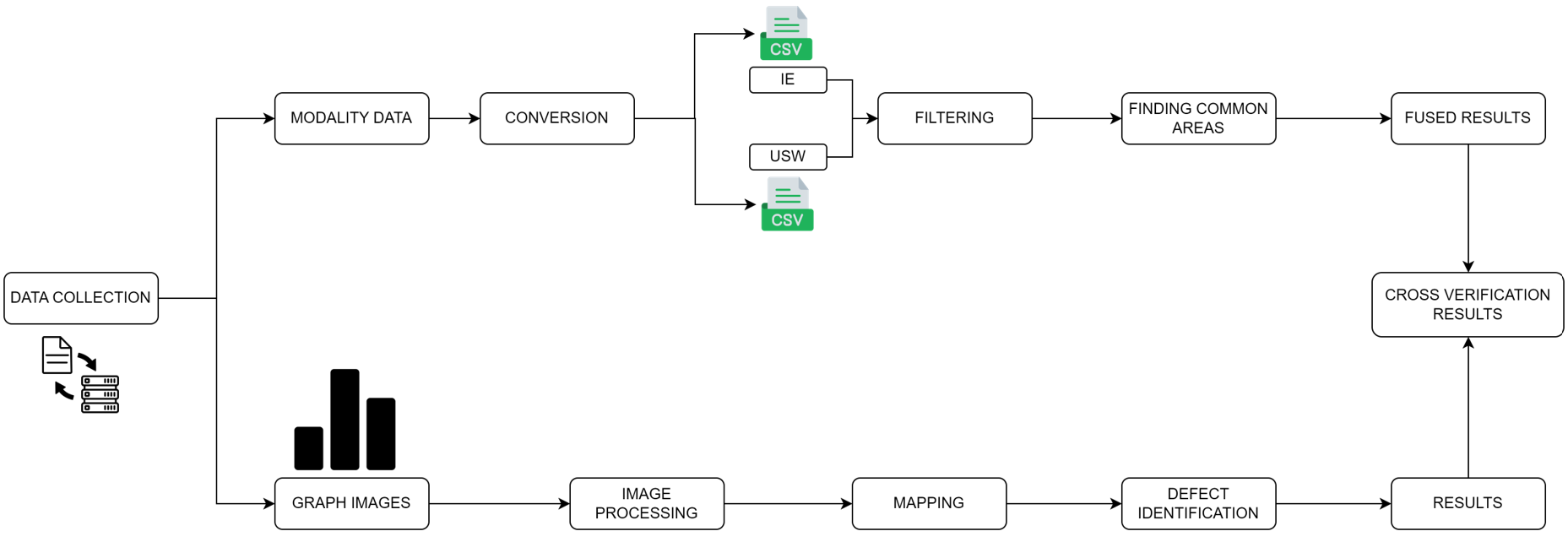}
    \caption{The Process Flow of the Framework}
    \label{fig:framework}
\end{figure*}

\subsection{Multimodal Data Fusion in SHM}

Multimodal data fusion has emerged as a solution to overcome the limitations of single-modality NDE techniques by integrating data from multiple sources. Fusing data from IE and USW enhances defect detection by combining elasticity and internal flaw detection capabilities. Research in SHM demonstrates that multimodal fusion improves the accuracy of defect localization and characterization, leveraging the strengths of each modality to create a more detailed and reliable diagnostic profile of bridge conditions \cite{b13, b10}. Spatial analysis techniques such as ASA \cite{b19} have been implemented in fusion-based approaches to map defects with greater spatial accuracy. Studies by Scherr and Grosse \cite{b14} and Zhang et al. \cite{b2} show that fusion-based defect mapping enables precise delineation of problematic areas, facilitating targeted inspections and maintenance. However, challenges remain in multimodal fusion, particularly with data harmonization and spatial alignment across modalities. Achieving seamless data integration and maintaining consistency across modalities are crucial for the reliability of fusion-based SHM frameworks \cite{b7, b6}.

\subsection{Cross-Verification with Image Processing }

Cross-verification through image processing has proven effective in enhancing the reliability of defect detection by validating fusion results with visual data. Techniques such as contour and gradient analysis are utilized to identify visually distinct defect patterns, often observed in contour maps or heat maps. Studies by Pozzer et al. \cite{b18} and Ichi et al.  \cite{b4} highlight the importance of aligning visual data points with fusion-based defect locations as a secondary validation layer. This alignment reduces false positives and provides inspectors with visual confirmation of defect-prone areas, as further supported by the Transportation Forum \cite{b15} and the Federal Highway Administration \cite{b13}.

\subsection{Integrating Multimodal Fusion and Cross-Verification with Image Processing}

Recent advancements in SHM have shown the potential for combining multimodal data fusion with cross-verification through image processing to achieve a comprehensive assessment of structural health. Integrating visual and quantitative data offers a holistic perspective, addressing the limitations of single-modality approaches and reducing dependency on subjective interpretations \cite{b15, b10}. Scherr et al. \cite{b14} and Momtaz et al. \cite{b1} indicate that cross-verifying defect locations identified through multimodal fusion with image-based methods improves diagnostic accuracy and reliability.

Building on these findings, this research refines and implements a multimodal framework combining fusion and image processing to enable robust and scalable SHM \cite{b10}.

\section{Methodology} 

This section outlines the proposed methodology, detailing the sequential processes of data acquisition, integrated techniques for multimodal defect detection, data fusion and cross-verification to ensure a comprehensive structural assessment.

\subsection{The Framework}

The flowchart of the framework shown in Figure \ref{fig:framework} outlines the sequence and interactions of processing data and images to identify defects and generate results. It involves data acquisition, preprocessing, fusion, and cross-verification for defect detection. Specifically, IE and USW data from InfoBridge is interpolated and filtered to identify defect-prone regions \cite{b8, b10}. ASA is then employed to fuse these areas, generating a unified defect map that accurately identifies zones of concern  \cite{b19}. Image processing validates these defect regions using HSV (Hue, Saturation, Value) filtering and with the adaptive bounding boxes on contour graph images. This cross-verification approach ensures robust defect detection, enabling a comprehensive bridge health assessment and facilitating proactive structural monitoring
\cite{b12, b21}.

\vspace{-1mm}
\subsection{Dataset Description}

The dataset utilized in this study comprises IE and USW data sourced from InfoBridge, a publicly accessible platform provided by the FHWA, with spatial and material property measurements \cite{b13}. Specifically, data related to six bridges were extracted for this analysis and results related to a single bridge are chosen to be interpreted here. This dataset provides comprehensive defect detection features derived from NDE techniques \cite{b12}. Contour plots (graph images) which are also sourced from InfoBridge are employed for cross-verification and visualization \cite{b5}.

\subsubsection{ IE Data Features}

IE is utilized to detect internal flaws within concrete structures by analyzing the frequency response generated by stress waves. This method effectively identifies critical subsurface defects, including voids, cracks, and delamination \cite{b7}. The key IE features include spatial coordinates and a voltage array. The X and Y coordinates indicate the location of each measurement on the bridge deck. The voltage array is a sequence of voltage values representing the captured wave response, used to derive frequency information critical for locating anomalies \cite{b7}.

\subsubsection{USW Data Features}

USW measures the modulus of elasticity of concrete to evaluate surface stiffness and material degradation. This technique detects areas of reduced stiffness that may indicate deterioration, debonding, or structural weaknesses \cite{b9}. The key USW features include spatial coordinates and elasticity modulus data. The X and Y coordinates denote sensor placement on the bridge deck.

Elasticity modulus data, consisting of two voltage arrays (input and received signals), are used to derive phase velocity and elasticity modulus, identifying regions with compromised structural integrity \cite{b9,b10}.

\subsubsection{IE and USW Contour Maps from InfoBridge}

The contour maps of IE and USW data from InfoBridge provide insights into surface conditions and structural defects, including delamination and debonding \cite{b11, b14}. These graphical representations are crucial for cross-verification, as image processing techniques validate defect-prone areas identified through multimodal data fusion. By aligning visual evidence from contour maps with findings from data fusion, the reliability of defect detection is significantly enhanced \cite{b17, b18}.

\subsection{Data Conversion and Feature Calculation}

In this phase, the IE and USW data initially collected in XML format are converted to CSV files to facilitate readability, structured processing and analysis \cite{b13}.

\paragraph{IE Conversion and Frequency Calculation} 

The IE data consist of voltage arrays that represent captured wave responses. Using Fast Fourier Transform (FFT), these signals are converted from the time domain to the frequency domain \cite{b23}. FFT identifies the frequency with the maximum amplitude, referred to as the peak frequency. This frequency corresponds to the slab's thickness resonance, enabling the detection of structural anomalies such as voids or delamination \cite{b7}, \cite{b21}.

\paragraph{USW Elasticity Modulus Calculation} 

The USW data comprises voltage arrays representing captured surface wave responses, including input and received signals from two sensors, which are cross-correlated to compute the time delay between the signals and then used to estimate the Rayleigh wave velocity\cite{b10}. The Rayleigh velocity, accounting for material properties such as Poisson's ratio and concrete density, is converted into the shear wave velocity. Using the shear wave velocity, the elasticity modulus is calculated based on the material's dynamic properties. The derived elasticity modulus, a critical parameter, enables the assessment of the material's stiffness and identification of structural anomalies, including changes in elasticity due to damage or deterioration\cite{b10}.

\vspace{-1mm}
\subsection{Defect Identification through Multimodal Data Fusion}

Defect detection using multimodal data fusion involves two essential steps: modality-specific filtering and identifying shared regions between the modalities.

\subsubsection{Defect Filtering}

In the Filtering stage, to isolate high-risk zones where structural defects are most likely present, empirically determined thresholds are applied to the IE and USW datasets:
\paragraph{IE Defect Filtering} The peak frequency threshold for the defect detection is determined by applying k-means clustering\cite{b27} to partition the frequency data into three clusters based on their values. The maximum frequency of the cluster with the lowest frequency values is then selected as the threshold and applied in subsequent analyses. Frequencies below this threshold are indicative of potential subsurface anomalies, such as voids or fractures, due to reduced resonance caused by material discontinuities \cite{b6}, \cite{b26}.

\paragraph{USW Defect Filtering} Similarly, the elasticity modulus threshold for defect detection is determined using k-means clustering to identify regions with significant property variations. The minimum cluster center value is selected as the threshold and applied in subsequent analyses.  Lower modulus values suggest material degradation, debonding, or stiffness loss, which are indicative of structural weaknesses \cite{b5}, \cite{b9}, \cite{b26}.

By applying these thresholds, the analysis focuses on defect-prone regions, excluding structurally sound areas and enhancing the accuracy of defect detection.

\vspace{2mm}

\subsubsection{Finding Common Areas}

To identify common spatial regions between the IE and USW datasets, ASA is employed as a computational geometry technique. ASA extends the concept of convex hulls to flexibly capture the boundaries of irregular and concave shapes within the datasets. This enables the delineation of defect-prone regions where data points are clustered, providing a reliable approach to localize anomalies.

Given a set of points \( P = \{p_1, p_2, \dots, p_n\} \) in a two-dimensional space, ASA constructs alpha shapes (\(\alpha\)-shapes) based on a parameter \(\alpha\) that determines the tightness of the boundaries. Larger \(\alpha\) values create convex hull-like boundaries, while smaller \(\alpha\) values reveal more nuanced concavities and clusters. This allows fine-grained identification of defect-prone zones by adjusting the level of detail to match the data distribution \cite{b19}.

Independent alpha shapes (\(\alpha_{IE}\) and \(\alpha_{USW}\)) are generated for the IE and USW datasets, respectively. The common defective regions are identified by computing the intersection of these alpha shapes:
\[
\alpha_{\text{common}} = \alpha_{IE} \cap \alpha_{USW}
\]
This intersection identifies spatial regions where both datasets indicate potential structural anomalies. These common areas are further refined by overlaying unified lane boundaries, providing spatial context and supporting the fusion of defect-prone zones.

The process generates a fused dataset comprising defect-prone regions identified by both modalities. The fused dataset serves as a key intermediate step, enabling a detailed spatial map of potential anomalies to inform subsequent visual cross-verification and analysis \cite{b3, b10, b19}.

\vspace{-1mm}

\subsection{Cross Verification through Multimodal Data Fusion}

Cross verification through multimodal data fusion include three key stages: source image processing, defection bounding box detecting, and fusion of defect data for verification.

\vspace{-1mm}
\subsubsection{Source Image Processing} 

The input images for defect identification are sourced from InfoBridge. These images provide a visual representation of defect-prone areas, highlighting spatial variations in Peak Frequency for IE and Elasticity Modulus for USW \cite{b7}, \cite{b12}. The examples of IE and USW images from FHWA are shown in Figures \ref{fig:fhwa_IE} and \ref{fig:fhwa_USW} of the Appendix, respectively.

Image processing is employed to extract potential defect regions from the input graphs. This stage involves several steps to accurately identify and map defective areas:

\paragraph{Color and Gradient Masking} Using OpenCV \cite{b20}, images are processed to detect red regions indicative of defects using predefined HSV color thresholds, identifying red and gradient regions extending into yellow hues. This masking ensures comprehensive defect detection that might not be captured by a narrow color range \cite{b18}.

\paragraph {Adaptive Morphological Operations} To refine defect regions and suppress noise, morphological operations are applied adaptively. The kernel size and iteration count are dynamically adjusted based on edge density, calculated using the Canny edge detector. This enhances the resolution of defect-prone areas while eliminating minor noise, improving detection accuracy \cite{b18,b20}.

\subsubsection{Defection Identification}

\paragraph{Contour Detection} Contours are extracted from the composite masks, with bounding boxes drawn around significant regions. Bounding boxes with an area below a predefined threshold (40 pixels) are ignored to filter out noise, ensuring that only meaningful defect regions are analyzed \cite{b4,b19}.

\paragraph{Mapping}
To accurately represent the spatial context of detected defects, pixel coordinates from the bounding boxes are mapped to real-world data coordinates using axis ranges derived from the corresponding CSV files. A coordinate mapping function accounts for image cropping, scaling, and orientation. This ensures a seamless transition from image-based detection to structured spatial analysis \cite{b20}.

\subsubsection{Fusion of Defect Data for Verification}

This step validates the consistency and reliability of defect detection results by integrating fused defect data with image-based analysis. Fused defect locations are overlaid onto processed contour images to ensure alignment between defect-prone areas identified through multimodal data fusion and visible patterns on contour maps. Visual indicators, such as color gradients or distinct intensity changes, provide corroborative evidence that supports the accuracy of data-driven findings \cite{b18}. Mapping fused defect points to corresponding locations on contour images involves pixel-to-data transformation techniques, maintaining spatial consistency throughout \cite{b20}. By integrating data with image analysis, this cross-verification step acts as a validation layer, reducing false positives and enhancing the reliability of defect detection outcomes \cite{b21}.

\begin{figure}[t]
    \centering
    \includegraphics[width=0.45\textwidth]
    {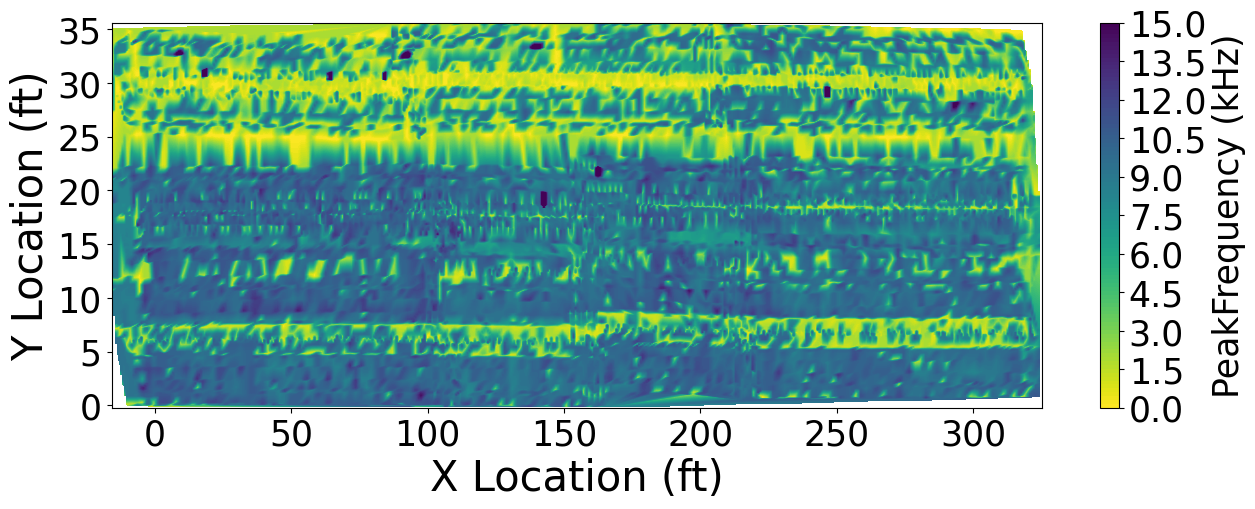}
    \caption{Contour plot illustrating the spatial distribution of IE data}
    \label{fig:IE_Contour_plot}
\end{figure}

\begin{figure}[t]
    \centering
    \includegraphics[width=0.45\textwidth]
    {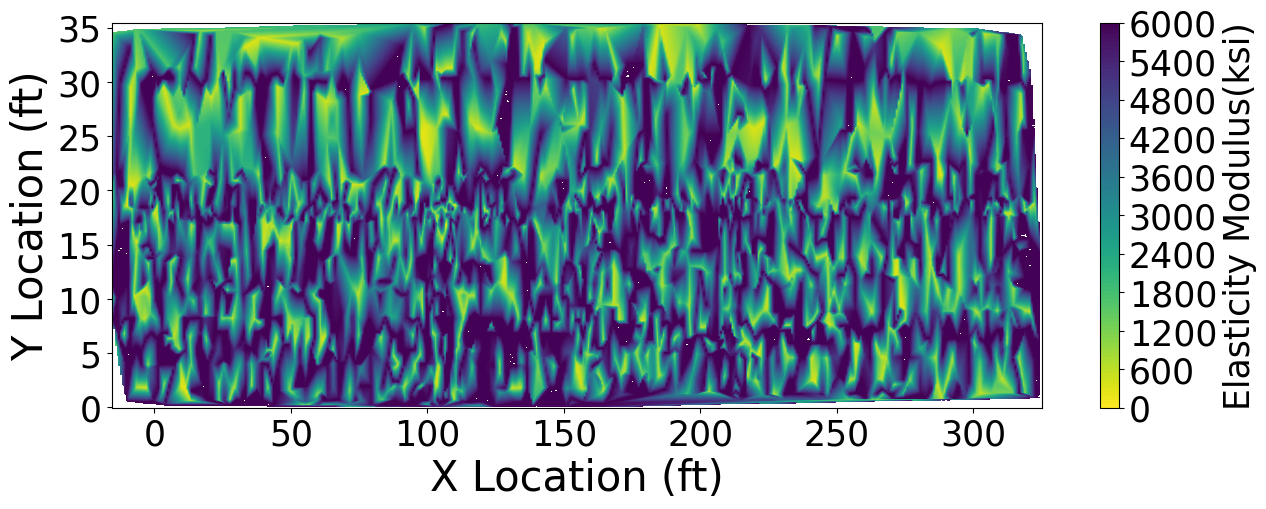}
    \caption{Contour plot illustrating the spatial distribution of USW data.}
    \label{fig:USW_contour}
\end{figure}

\section{Experimental Results and Discussion} 
This pilot study is applied to six bridges in Mississippi using data from InfoBridge, an open-source platform by the Federal Highway Administration (FHWA), to evaluate the feasibility and performance of the framework. However, for this paper, results and analyses are focused on one representative bridge to illustrate the framework's application and effectiveness. 
\subsection{Experimental Setup}
\paragraph{Platform and Tools} 
The implementation was conducted in python, using numpy, pandas, matplotlib, and scikit-learn for data processing and analysis. OpenCV was employed for image processing to detect defect regions. The alpha shape library defined spatial defect boundaries. The dataset used in this study corresponds to the I-220 over John R. Lynch St in Jackson, Mississippi. This bridge, identified by Structure Number 11002200250005B and LTBP Bridge Number 28 - 000007 as per the InfoBridge records, was chosen as a representative structure for the experimental results presented in this paper.

\paragraph{Hardware} The experiments were conducted on high-performance Central Processing Units (CPU's), capable of managing large datasets and computationally intensive tasks. This configuration supported Fourier Transform analysis for frequency-based defect detection, alongside comprehensive spatial processing for multimodal defect characterization.
\vspace{-1mm}
\subsection{Data Conversion and Feature Calculation}

\subsubsection{IE Frequency Calculation}
The contour plot in Figure \ref{fig:IE_Contour_plot} visualizes the spatial distribution of Peak Frequency (kHz) derived from IE data, represented across the bridge. The X and Y axes denote the spatial coordinates, while the color scale corresponds to the measured frequency values. High-frequency regions (blue–purple, $>10$ kHz) reflect areas with minimal subsurface irregularities, whereas low-frequency regions (yellow–green, $<5$ kHz) potentially indicate delaminations, voids, or other structural anomalies \cite{b25, b26}. These insights contribute to identifying defect-prone zones, offering a quantitative basis for defect detection and supporting the multimodal fusion framework described in this study.

\begin{figure} [t]
    \centering
    \includegraphics[width=0.45\textwidth]{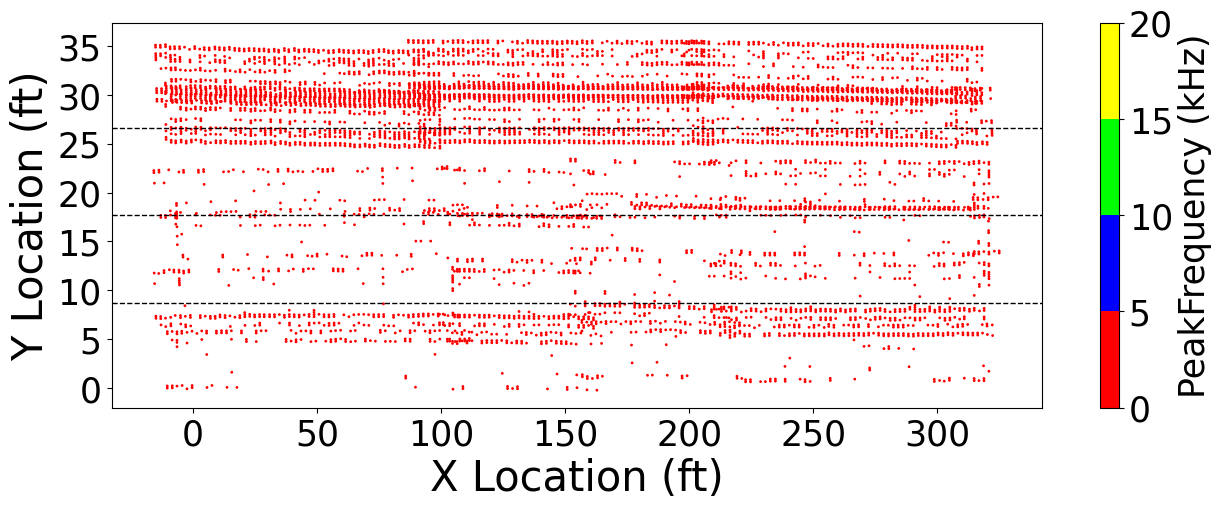}
    \caption{Scatter plot of IE defective data post filtering.}
    \label{fig:IE_Scatter_Plot}
\end{figure}

\begin{figure} [t]
    \centering
    \includegraphics[width=0.45\textwidth]
    {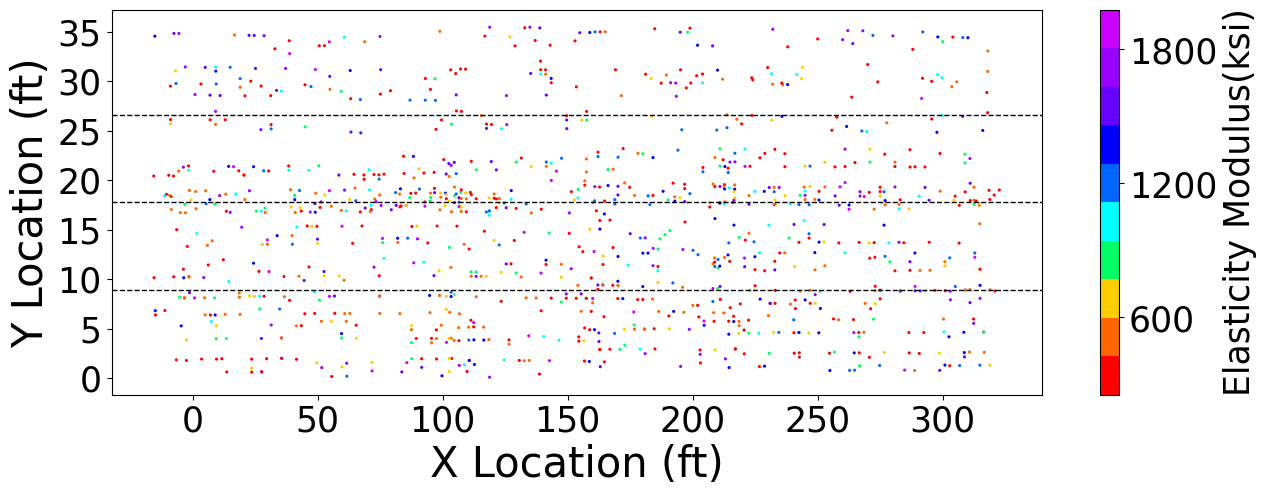}
    \caption{Scatter plot of USW defective data post filtering.}
    \label{fig:USW_Scatter_Plot}
\end{figure}

\subsubsection{USW Elasticity Modulus Calculation}

The contour plot in Figure \ref{fig:USW_contour} illustrates the spatial distribution of Elasticity Modulus (ksi) derived from USW measurements, mapped across the bridge. The X and Y axes represent the spatial coordinates, while the color scale denotes the elasticity modulus values. High modulus regions (blue–purple, $>4000$ ksi) signify areas of greater material stiffness, indicating structurally sound zones. Conversely, low modulus regions (yellow–green, $ <2000 $ ksi) suggest potential material degradation, such as debonding or reduced stiffness, which may compromise structural integrity \cite{b25, b26}. The interpolation of the dataset reveals spatial variations and patterns that may correspond to localized subsurface anomalies. These findings contribute to identifying defect-prone regions and serve as a critical input for multimodal data fusion.

\vspace{-1mm}

\subsection{Defect Filtering}

\subsubsection{IE Defect Filtering}
The IE peak frequency threshold of 4.31kHZ for defect filtering is established using k-means clustering with $K$ = 3\cite{b27}. 
The scatter plot in Figure \ref{fig:IE_Scatter_Plot} presents the spatial distribution of Peak Frequency (kHz) values below the threshold captured through IE data. The color-coded markers highlight the frequency ranges, with red indicating zones of potential structural concern such as delaminations or voids \cite{b25, b26}. Horizontal dashed lines divide the vertical axis into uniform sections, corresponding to lanes of the bridge. The prevalence of red markers along these sections suggests concentrated defect-prone zones, offering clear spatial context for further investigation.

\subsubsection{USW Defect Filtering} The USW elasticity modulus threshold of 2012 ksi for defect filtering is established using k-means clustering with $K$ = 3\cite{b27}.
The scatter plot in Figure \ref{fig:USW_Scatter_Plot} depicts the spatial distribution of Elasticity Modulus (ksi) from USW measurements, filtered to values below the threshold to highlight defect-prone regions\cite{b25,b26}.The X and Y axes indicate spatial coordinates, while the color scale reflects stiffness variations across the bridge. Lower Elasticity Modulus values (e.g., red and orange) suggest potential structural issues like material degradation or stiffness loss. Horizontal dashed lines divide the vertical axis into uniform sections, corresponding to the lanes of the bridge, aiding in localized investigations and supporting structural health monitoring through multimodal fusion analysis.

\subsection{Finding Common Areas}

\begin{figure} [t] 
\centering 
\includegraphics[width=0.45\textwidth]
{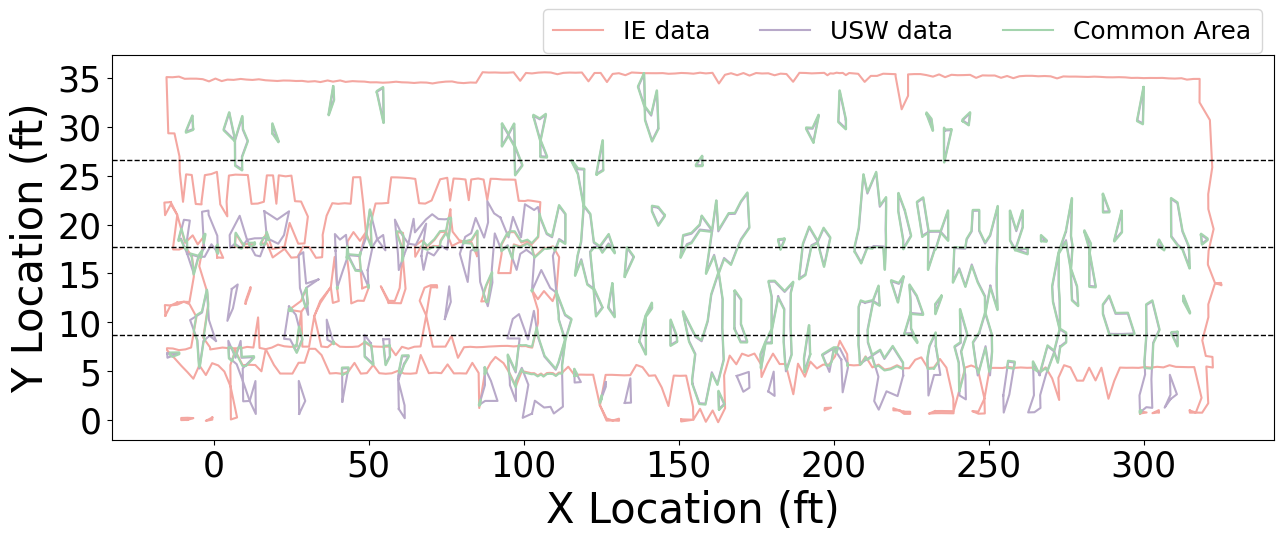}
\caption{Visualization of common areas from IE and USW identified by ASA} 
\label{fig:Alpha_Shape_Intersection} 
\end{figure}

The plot in Figure \ref{fig:Alpha_Shape_Intersection} illustrates the alpha shape boundaries for the IE (red) and USW (purple) datasets, with the green-shaded regions representing the intersecting common areas where both modalities indicate potential defects. The X and Y axes correspond to spatial coordinates, while the alpha shapes ($\alpha = 0.5$) delineate concave boundaries of defect-prone regions. The intersecting areas enhance defect localization by corroborating anomalies detected in both datasets. Horizontal dashed lines indicate lane boundaries, aiding in the spatial interpretation of structural anomalies for targeted health assessments of the bridge.

\subsection{Data Fusion of IE and USW Signals}

The fused results identify unique defective locations by integrating data from IE and USW techniques. Low values in both Peak Frequency and Elasticity Modulus represent potential anomalies, enabling the detection of defect-prone regions across both modalities. By merging spatial coordinates from these datasets, the fusion framework highlights all unique defect sites, offering a comprehensive view of defective regions. This integration leverages multimodal data to enhance the reliability of defect detection and localization \cite{b2, b5, b10}.

\begin{figure} [h]
    \centering\includegraphics[width=0.45\textwidth]{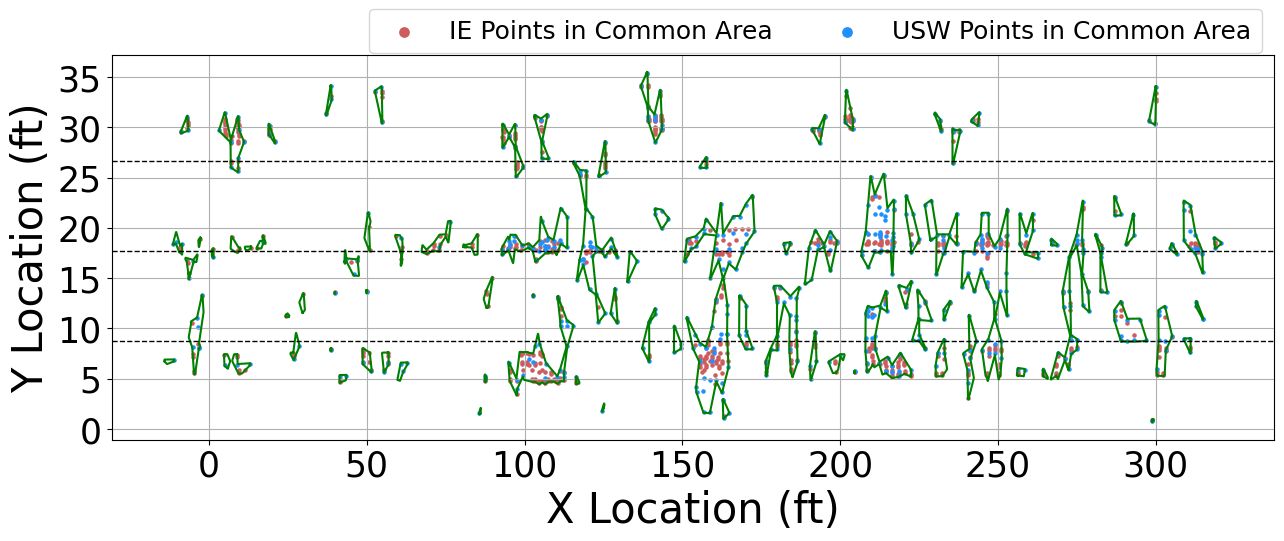}
    \caption{Common defective areas of IE and USW identified through Fusion}
    \label{fig:IE_USW_Fusion}
\end{figure}

The scatter plot shown in Figure \ref{fig:IE_USW_Fusion} highlights the intersection of defect-prone regions identified using IE and USW data, visualized within the common area of analysis. Points from IE data are represented in red, while points from USW data are depicted in blue, both confined within the green-outlined common area derived through alpha shape analysis \cite{b19}. The spatial overlap of these data points signifies regions with high defect likelihood, corroborated across both modalities. Such intersections reinforce the reliability of the multimodal fusion framework by cross-validating defect zones \cite{b3, b10}.

\subsection{Defect Identification through Image Processing}

\subsubsection{Image Processing}

In this step, image data is processed using HSV-based color thresholds to isolate potential defect regions, followed by adaptive morphological operations to refine defect boundaries and reduce noise. 

The plot in Appendix Figure \ref{fig:IE_axis} represents the spatial distribution of frequency data from IE measurements, mapped to real-world coordinates using CSV-based axis ranges. The X and Y axes denote the physical bridge locations, accurately scaled from pixel data. Color mapping includes blue for structurally sound regions and green-to-yellow gradient for potential subsurface anomalies, ensuring precise defect localization.

Similarly, the plot in Appendix Figure \ref{fig:USW_axis}  shows the Elasticity Modulus distribution derived from USW data, mapped to real-world coordinates using corrected axis ranges. The X and Y axes now align with physical bridge dimensions. Color mapping includes blue for structurally sound regions and green-to-yellow for areas of potential subsurface anomalies, ensuring precise defect localization.

\subsubsection{Defect Detection}

Defect detection isolates potential anomalies by segmenting regions using HSV-based color thresholds, identifying red-gradient zones for low elasticity modulus (USW) and low-frequency values (IE) \cite{b18}. Contours are detected, refined with adaptive morphological operations, and bounded by boxes excluding noise below a threshold area \cite{b20}. Then they are mapped to real-world coordinates, aligning visually detected defects with spatial data, ensuring consistency with multimodal fusion results \cite{b4}.

\begin{figure} [h] \centering \includegraphics[width=0.4\textwidth]{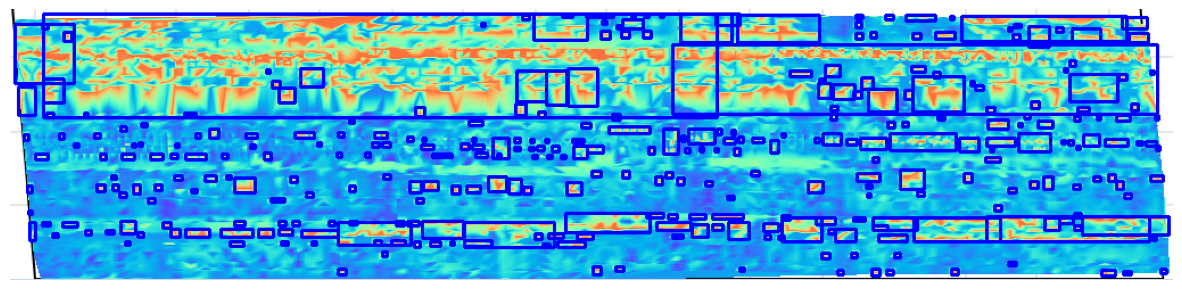} \caption{Defective regions of IE: Bounding boxes indicate frequency inconsistencies} \label{fig: IE_BoundingBoxes} 
\end{figure}

\begin{figure} [h] \centering \includegraphics[width=0.4\textwidth]{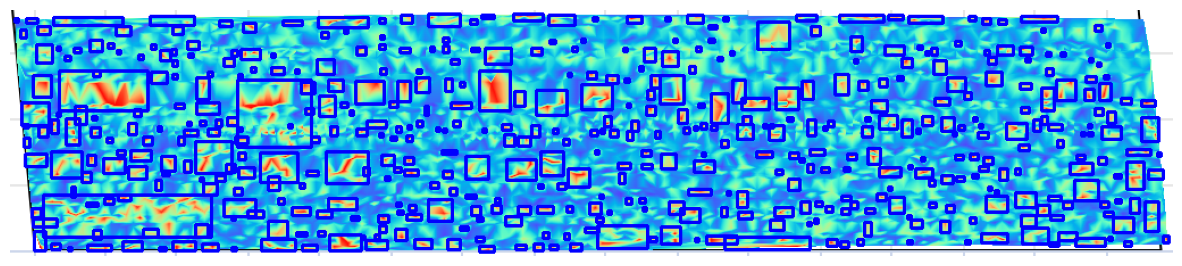} \caption{Defective regions of USW: Bounding boxes indicate modulus inconsistencies} 
\label{fig: USW_BoundingBoxes} 
\end{figure}

The IE contour plot in Figure \ref{fig: IE_BoundingBoxes} visualizes variations in Peak Frequency (kHz) across the structure, with red-yellow regions highlighting potential defects such as delaminations or voids, and blue regions representing structurally sound areas. Superimposed bounding boxes, generated through adaptive morphological analysis and contour detection, accurately enclose the defect-prone areas. These contours provide a detailed spatial representation of frequency anomalies, enabling precise localization of subsurface structural issues.

The USW contour plot in Figure \ref{fig: USW_BoundingBoxes} displays the distribution of Elasticity Modulus (ksi) across the structure, with red-yellow regions indicating areas of potential material degradation or stiffness loss, and blue regions marking sound zones. Bounding boxes derived from contour detection and adaptive processing highlight defect-prone areas with precision. The contours provide detailed insights into the spatial patterns of stiffness anomalies, facilitating accurate identification and localization of structural weaknesses.

\subsection{Cross-Verification}

Visual cross-verification validates fused results by overlaying defect data points onto contour images from IE and USW. Processed images highlight defect-prone areas, aligning spatially with fused data for qualitative assessment. Bounding boxes and adaptive morphology techniques refine defect boundaries, ensuring consistency between multimodal fusion results and image-based findings \cite{b4, b15, b18}.

\begin{figure} [t]
    \centering
    \includegraphics[width=0.45\textwidth]{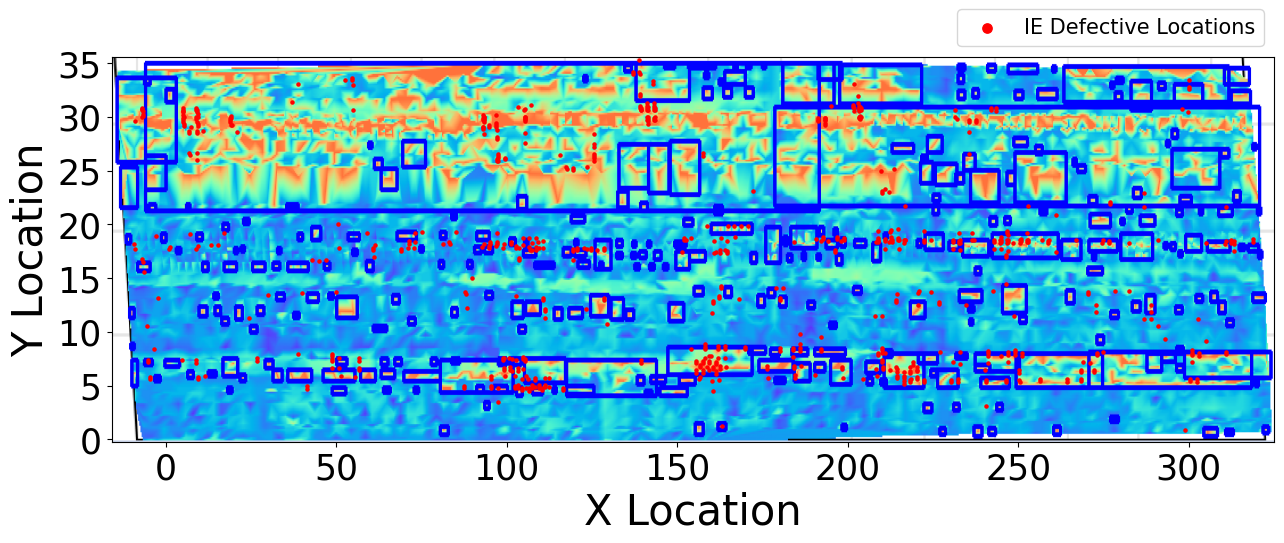}
    \caption{IE defective locations overlaid on the processed contour plots from InfoBridge}
    \label{fig:IE_newaxis}
\end{figure}

The contour map in Figure \ref{fig:IE_newaxis} overlays fusion defect points (red markers) onto the cropped IE data visualization, which includes bounding boxes (blue rectangles) highlighting localized anomalies. These bounding boxes represent low-frequency zones identified as potential delaminations or voids based on IE measurements. The red markers from the fusion process align with the IE anomalies, demonstrating consistency between data fusion results and contour-based defect localization. This combined representation effectively validates the reliability of bounding box detection and the multimodal fusion approach for accurately identifying structural defects.

\begin{figure} [t]
    \centering
    \includegraphics[width=0.45\textwidth]{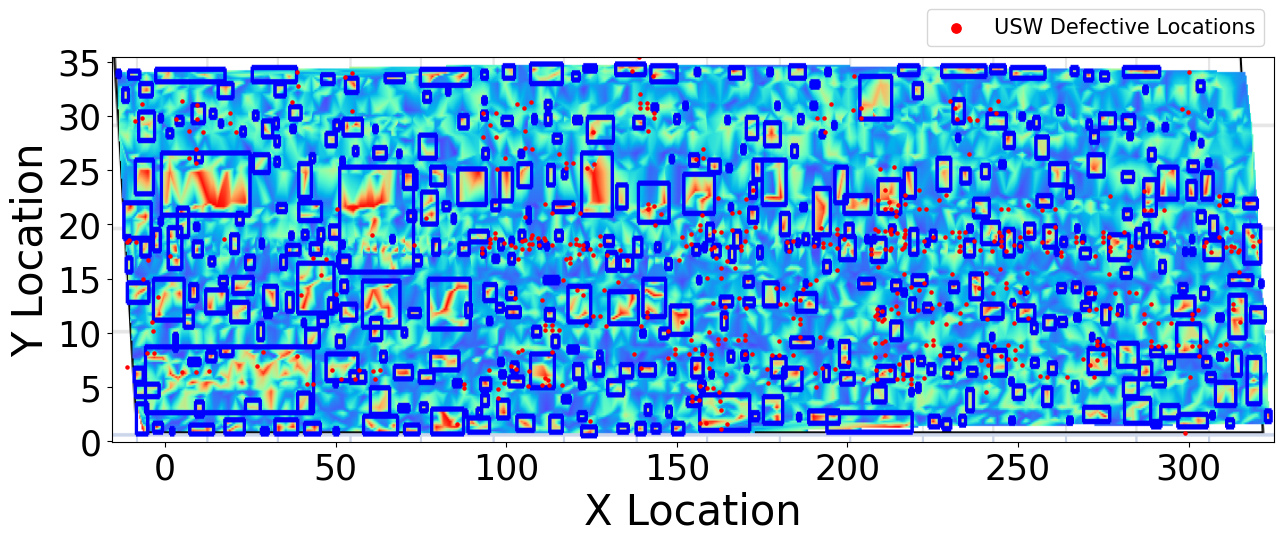}
    \caption{USW defective locations overlaid on the processed contour plots from InfoBridge}
    \label{fig:USW_newaxis}
\end{figure}

The contour plot in Figure \ref{fig:USW_newaxis} overlays fusion defect points (red markers) onto the cropped USW visualization, featuring bounding boxes (blue rectangles) that indicate low elasticity modulus regions associated with stiffness loss or material degradation. The red markers derived from multimodal data fusion align with the USW-identified defect zones, ensuring consistency between fusion results and contour-based anomalies. This visualization effectively consolidates USW anomalies, providing precise defect localization and robust cross-validation of structural weaknesses.

The combined evaluation metrics highlight the overall performance of the framework, demonstrating its capability to reliably detect structural defects by integrating multimodal data fusion and cross-verification techniques. Using micro-averaged metrics, which aggregate all true positives, false positives, and false negatives across both IE and USW datasets, the framework achieved a precision of 0.75, recall of 0.92, and F1-score of 0.83. Micro-averaging provides a global measure of performance by treating each defect point equally across datasets, ensuring a fair assessment of the system's ability to localize defects accurately \cite{b3, b21}. These results validate the robustness of the proposed approach \cite{b13}, ensuring high accuracy in defect detection while minimizing false positives \cite{b12} and delivering comprehensive structural assessments \cite{b15}.


\section{Conclusions and Future Work}

This pilot study presents an efficient framework for detecting bridge structural anomalies through the fusion of multimodal data, specifically integrating IE and USW measurements. Utilizing advanced geospatial analysis techniques such as alpha shapes and threshold-based filtering, the framework effectively identifies defect-prone regions. The fusion approach offers a comprehensive view of structural anomalies, while image-based cross-verification ensures high reliability by validating defect locations against visual data. This scalable methodology provides a significant tool for efficient bridge health assessment, with important implications for proactive structural monitoring and maintenance.

Future work aims to enhance the framework by incorporating real-time SHM data streams for continuous monitoring and predictive maintenance, thereby improving defect detection responsiveness and infrastructure safety. Expanding sensing modalities beyond IE and USW could address more complex anomalies, while adaptive image processing algorithms may enhance cross-verification accuracy and reduce false positives. These advancements would optimize resource allocation in critical infrastructure management, reinforcing defect detection reliability and offering a scalable and efficient toolkit for sustainable infrastructure management.

\section*{Acknowledgment}

The authors would like to express their sincere gratitude to the FHWA for their invaluable support and access to data, tools, and resources that were instrumental in the completion of this research. The open-source data and platforms provided by FHWA significantly broadened the scope and depth of our study on bridge defect detection and analysis. We are particularly grateful for the insights and resources that enabled us to implement and validate our multimodal framework, contributing to the advancement of structural health monitoring. This research would not have been possible without FHWA’s commitment to innovation and collaboration in infrastructure research.

\newpage
\section{Appendix}

\begin{figure}[!htb]
    \centering
    \begin{tabular}{cc}
        \begin{minipage}{0.45\textwidth}
            \centering
            \includegraphics[width=\linewidth]{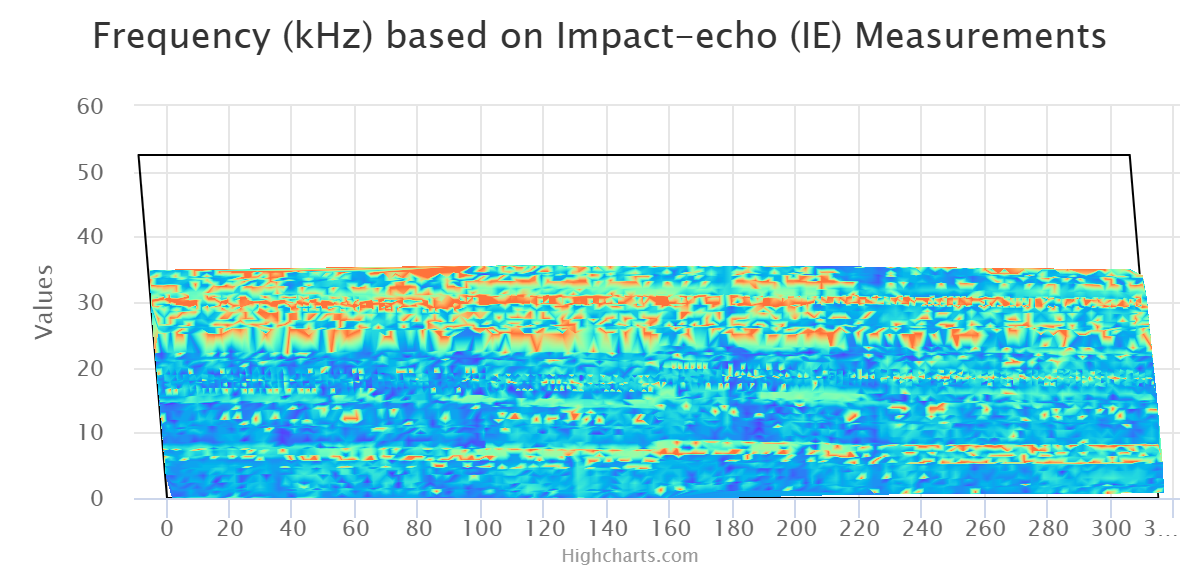}
            \captionof{figure}{IE Contour Image. Source: FHWA \cite{b24}.}
            \label{fig:fhwa_IE}
            
            \vspace{0.5cm} 
            
            \includegraphics[width=\linewidth]{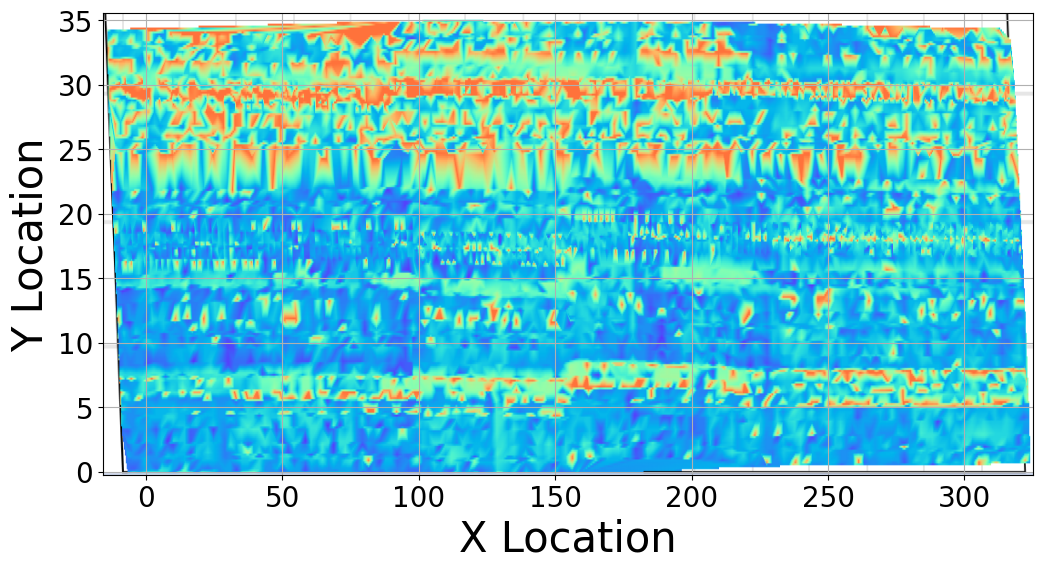}
            \captionof{figure}{IE Contour plot with mapped new axes.}
            \label{fig:IE_axis}
        \end{minipage} &
        
        
        \begin{minipage}{0.45\textwidth}
            \centering
            \includegraphics[width=\linewidth]{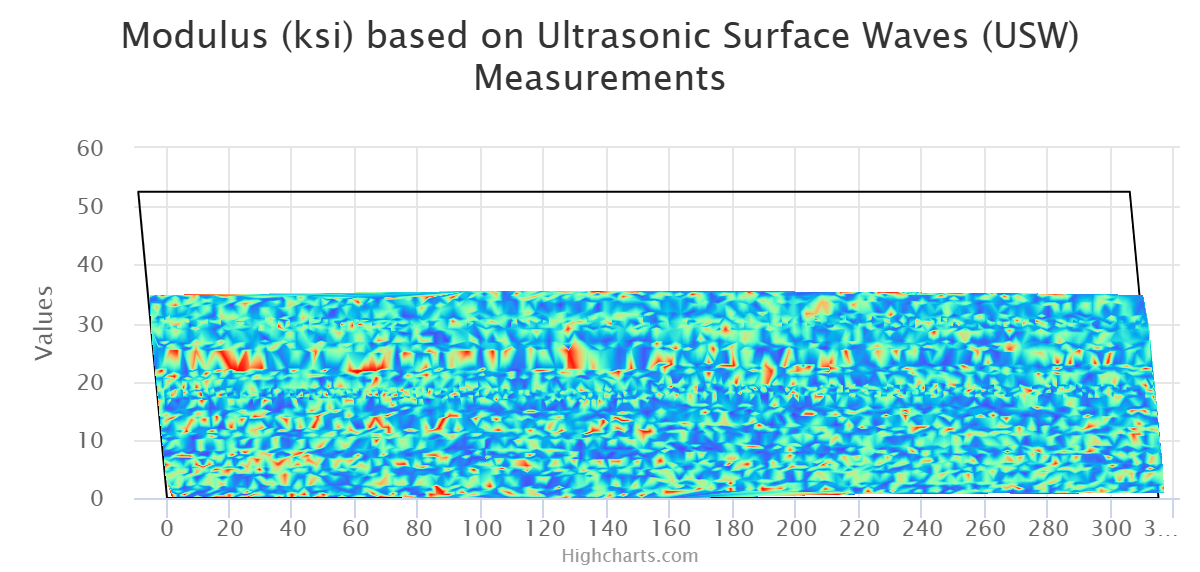}
            \captionof{figure}{USW Contour Image. Source: FHWA \cite{b24}.}
            \label{fig:fhwa_USW}
            
            \vspace{0.5cm} 
            
            \includegraphics[width=\linewidth]{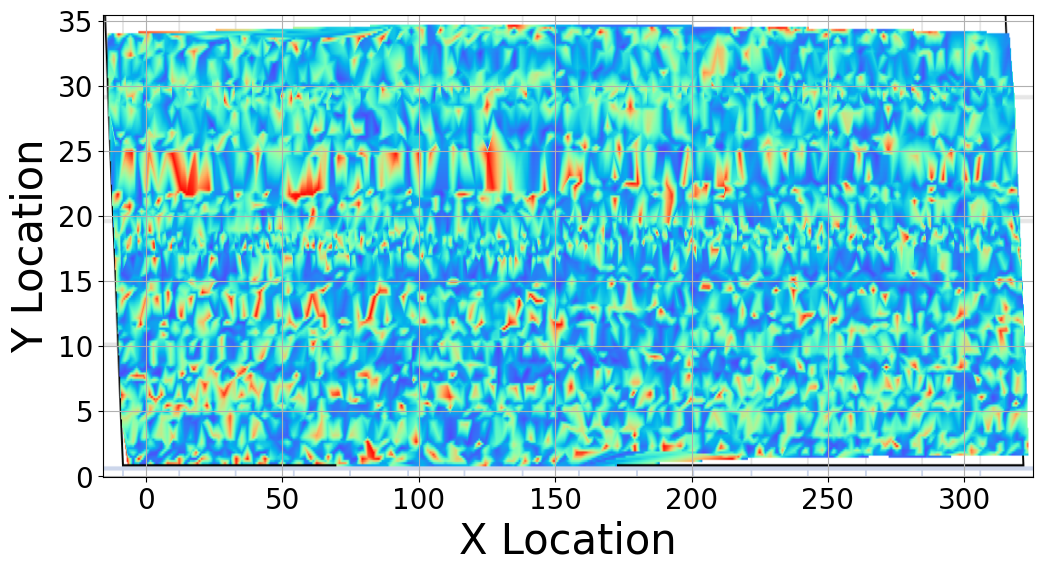}
            \captionof{figure}{USW Contour plot with mapped new axes.}
            \label{fig:USW_axis}
        \end{minipage}
    \end{tabular}
\end{figure}

\end{document}